\def\year{2015}
\documentclass[letterpaper]{article}
\usepackage{aaai}
\usepackage{times}
\usepackage{helvet}
\usepackage{courier}
\usepackage{graphicx,multirow}
\usepackage{amsmath}
\usepackage[all]{xy}
\usepackage{amssymb}
\usepackage{subfigure}
\usepackage{breqn}

\usepackage[font={sf,small}]{floatrow}
\usepackage{graphicx}
\usepackage{amsmath}
\usepackage{verbatim}
\usepackage{comment} 
\usepackage{algorithm}
\usepackage{algorithmic}
\usepackage{amsfonts}

\usepackage{multirow}
\usepackage{flushend}
\usepackage{xspace}
\usepackage{xcolor}
\usepackage{tabularx}
\newcolumntype{b}{X}
\newcolumntype{s}{>{\hsize=.23\hsize}X}
\newfloatcommand{capbtabbox}{table}[][\FBwidth]

\begin{document}
\title{Feature Based Task Recommendation in Crowdsourcing with Implicit Observations}
\author{Habibur Rahman\\
University of Texas at Arlington\\
habibur.rahman@mavs.uta.edu
\And
Lucas Joppa\\
Microsoft Research\\
ljoppa@microsoft.com
\And
Senjuti Basu Roy\\
New Jersey Institute of Technology \\
senjutib@njit.edu
}
\maketitle
\begin{abstract}
\begin{quote}
%With the growing popularity of crowdsourcing based applications, task recommendation to workers based on previous task completion  is an important research area. 
%Existing research in crowdsourcing has investigated how to recommend tasks to workers based on which task the workers have already completed, referred to as {\em implicit feedback}. 
We initiate the study of task recommendation problem for citizen science crowdsourcing platforms, where we leverage both {\em implicit feedback and explicit features of the tasks}. We assume that we are given a set of workers, a set of tasks, interactions (such as the number of times a worker has completed a particular task), and the presence of explicit features of each task (such as, task location). We intend to recommend tasks to the workers by exploiting implicit interactions, and the presence or absence of explicit features in the tasks. We present two alternative optimization problems,and propose respective solutions. We compare our solutions against multiple state-of-the-art techniques using a real world large citizen science dataset. 
\end{quote}
\end{abstract}

\section{Introduction}
Crowdsourcing platforms, such as Amazon's Mechanical Turk or Crowdflower,  have recently gained immense popularity due to  their elegant framework, where a task requester can get work done by numerous virtual workers for very low compensation. One common problem in these platforms is that workers have to suffer huge latency to find suitable tasks, which creates dissatisfaction and eventually leads to the abandonment of the platform. Task recommendation problems are studied in the crowdsourcing context,  where the objective is to recommend a set of tasks to  each worker such that these tasks are best suited for the workers~\cite{geiger2014personalized,yuen2012taskrec}. In this work, we aim at leveraging the task completion history of the workers (referred to as {\em implicit feedback}) and augment that with explicit task characteristics or features to recommend tasks to the workers. Our focus of investigation is limited to citizen science crowdsourcing applications where the importance of effective task recommendation is pivotal~\cite{xue2013improving}. We focus on the crowdsourcing of biodiversity observations, where volunteer visit sites, observe species, and report their findings via web applications.  Currently, a volunteer, upon identifying a species, uploads information into the server specifying the details of the identification. A common problem which frequently occurs in this scenario is {\em incorrect identification}. A reliable task recommender system can alleviate the problem. If we have historical data on how many tasks a volunteer has successfully performed and those observations are on what species and from which locations, then we can lower the risk in incorrect identification by asking volunteers to identify species they have prior experience with.

\section{Methodologies}
\label{sec:dataModel}
%\vspace{-0.05in}
{\bf Notations:} $W =\langle w_1,w_2,w_3 \dots w_{n_w} \rangle$ and $T= \langle t_1,t_2,t_3,\dots t_{n_t} \rangle$ represents the set of workers and tasks respectively. The relationship between workers and tasks is represented by matrix $C_{n_w\times n_t}$, where $c_{wi}$ represents the number of times worker $w$ has completed task $i$. The preference matrix $P$ is a boolean version of $C$, such that $p_{wi} =1$, if $c_{wi} \geq 1$, otherwise $p_{wi} =0$. %Each task can be described by the presence or absence of explicit task features, denoted as $F=\langle f_1,f_2,f_3,\dots f_{n_l}\rangle$.  
$Y_{n_t \times n_l}$ represents the explicit task feature matrix, where $y_{il} \in \{0,1\}$ denotes the absence or presence of feature $l$ for task $i$. Worker feature preference matrix is denoted as $X_{n_w \times n_l}$. Additionally, $U_{n_w \times n_f}$ and $V_{n_t \times n_f}$ are the two latent factor matrices, where $U$ is for the workers and $V$ is for the tasks. 
{\bf Formulation 1 - Feature Preference Model:} We assume that the reason that a particular worker has completed a particular task is because the worker has a hidden preference over the task features which we want to uncover. As an example, if locations are used as task features, we can learn the preference of workers for different location, which can be used for recommend new task to workers. Based on the  explicit knowledge of task feature matrix $Y$ and worker task completion matrix  we learn the preference of each worker in the feature space or $X$. Formally, we want to minimize the following objective function.
\begin{align}
M = \sum_{w,i}q_{wi}(p_{wi}-x_wy_i)^2+\lambda(\|X\|^2)\\ 
 X \geq 0\\
 q_{wi} = 1 + \alpha \times c_{wi}
\label{eqn:objective}
\end{align}
% 
% where the non-negativity constraint must be satisfied. Such that, 
% \begin{align*}
%   X \geq 0
% \end{align*}
% 
% and,
% 
% \begin{align*}
%  w(x) = 1 + \alpha \times c(x)
% \end{align*}
Here, $q_{wi}$ is designed such that, the weight of positive signals is amplified. $Q$ denotes the matrix representing the values of $q_{wi}$ for all workers and tasks. If a particular observation has high confidence the system will choose $x_w$ such that $x_wy_i$ becomes close to $1$. $\alpha$ is set to a positive value indicating the confidence for the positive signals over negative signals. By solving $M$, we get the solution for user vector, $x_w =  (Y^tQ^wY + \lambda I)^{-1}Y^tQ^wP_w$. Due to the non-negativity constratint of $X$, we solve the following optimization problem as $\|(Y^tQ^wY + \lambda I)x_w - Y^tQ^wP_w\|^2$. We refer to our algorithm as {\tt Feat-Based-NNLS} or Feature Based Non-Negative-Least Square.    

{\bf Formulation 2: Latent Factor Model:} We consider the following objective function for task recommendation -  
\begin{dmath}
\label{eqn:objective2}
  M = \sum_{w,i}q_{wi}(p_{wi}-u_wv_i)^2+\lambda(\|U\|^2+ \|V\|^2 - \sum_{i,i'}v_i^tv_i'Sim(i,i'))
\end{dmath}
Here, the goal is to find $U$ and $V$ such that it minimizes the error, where $\lambda$ is the regularization parameter. For any  new task, the predicted recommendation score is calculated by multiplying $U_w$ with $V_i$. To incorporate the task similarity into the latent factor based formulation, we add a penalty term in the equation. Our intuition is that if the similarity between any two tasks is high, then they should also be similar in the latent factor space. Our notion of task similarity is defined as $sim(t_i, t_j)$ = $\frac{1}{1 + e^{-Y_i^tY_j}}$. The analytical solution for $U$ and $V$ is given below.

\begin{align}
 \label{eqn:user_latent}
 u_w =  (V^tQ^uV + \lambda I)^{-1}V^tQ^wP_w
\end{align}
\begin{align}
v_i =  (U^tQ^iU + \lambda I)^{-1}(U^tQ^iP_i + \lambda * 0.5 * \sum_{i'=1}^{n_t} Sim(i,i')v_i') 
\end{align}
We solve the optimization problem by alternating and fixing $U$ and $V$. This method is referred to as Implicit Factorization with Task Similarity or {\tt IFTS}.  
\vspace{-0.1in}
\section{Experiments}
\label{sec:exp}
%{\bf poorly written - full of typos - inconsistent tense - i fixed a few - needs more work}
%We present our experimental evaluations next.
% First we describe the dataset, then we describe our baseline algorithms and evaluation metrics and later the evaulation summary. 
% \textbf{System:} Our development and test environment uses Python 2.7 on a linux Ubuntu 14.04 machine, with Intel Core i5 2.3 GHz processor and a 6-GB Ram. All numbers are presented as the average of three runs.

% \subsection{Dataset Descriptions}
% 
% \begin{figure}[ht]
% \centering
% \subfigure[{\tt Inat}]{
%    \includegraphics[height=2.5cm, width=3.9cm] {figures/workerTaskDist_Inat.pdf}
%      \label{fig:workerTaskDist_Inat}
%  }
%  \subfigure[{\tt Ebird}]{
%    \includegraphics[height=2.5cm, width=3.9cm] {figures/workerTaskDist_Ebird.pdf}
%     %{figures/synthetic/matlab/fig2SimTimeVsThroughput.pdf}
%    %\caption{(b)}
%     \label{fig:workerTaskDist_Ebird}
%  }
%  \label{fig:hypo}
% \caption{\small Worker Task Distribution}
% \end{figure}
% 

%\begin{figure}[ht]
%\centering
%\includegraphics[scale=0.5] {figures/workerTaskDist.pdf}
%\caption{\small {\bf Workers Per Task}}
%\label{fig:workerTaskDist}
%\end{figure} 

{\bf Dataset:} We collected data from a popular citizen science platform named {\em Ebird }\footnote{Ebird.org}. Ebird is a popular citizen science platform for bird observations. We crawled all the observations from year $2012$ and randomly choose a set of $5000$ workers for our experiments leading to $1767$ tasks with a total number of $2.5$ million observations. We used $294$ locations as task features.%Worker task distribution for this dataset is given in Figure~\ref{fig:workerTaskDist_Ebird} 
{\em Evaluation:} We evaluate our methods using a hold out test set. We randomly choose $90\%$ of our data as the training set and remaining $10\%$ as the test set which gives us the ground truth. All the results are an average of three runs. 

% 
% \noindent {\em Inat:} In this dataset we crawl the data from August 2013 to July 2015 of about $1.5$ million observations. Each tuple in this dataset is an observation in the form of $\langle$ Observer, Latitude, Longitude, Species, Date$\rangle$.  We consider $317000$ observations of type {\em Bird} in our Experiments. Each observer is a worker and each species as a task. Our final dataset contains $4444$ workers and $19764$ tasks, average number of tasks per worker is $65.23$. We extract the location information from latitude and longitude and use it as explicit task features. 
%{\em Ebird:} 
%{\bf this is not clear - what do you mean?}     

{\bf Implemented Baseline Algorithms:}
% i) {\tt Implict-ALS}: This algorithm is designed according to ~\cite{hu2008collaborative}. This algorithm uses alternating least square method to factorize Worker-Task completion matrix considering implicit feedback. 
i){\tt Implicit-ALS-Neg}: This algorithm is implemented according to~\cite{lin2014signals}.The algorithm uses alternating least square method considering negative signals. If a worker has not completed a task then the total number of times that task has been completed by other users is considered as the weight of the negative signal. 
ii) {\tt Feature-Based-Reg}: We assume that the task-feature matrix $V$ is given to us. We solve the regularized regression~\cite{wu2006learning} problem $(C_{ij} - x_iy_j)^2 + \lambda \|X\|^2 $ to find $X$.

%If a worker has not completed a task we consider the total numberz of times that task has been completed by other users as the weight of the negative signals. This similar technique is also used in the original paper.     

%The value of $w$ is set according to the equation~\ref{}. Here, $f(x)= \frac{1}{1 + e^{\alpha_1(log(x) + \beta_1)}}$ and $g(x)= \frac{1}{1 + e^{\alpha_2(log(x) + \beta_2)}}$. We choose $\alpha_1$ = , $\alpha_2$ = , $\beta_1$ = , $\beta_2$ =  

% 

%iv) {\tt Neighborhood-Based}: This is similar to the Neighborhood Based algorithm implemented in ~\cite{hu2008collaborative}. For all pair of tasks $t_1$ and $t_2$, we compute the similarity $sim_{t_1t_2} = \frac{c_{t1}^Tc_{t2}}{\|c_{t_1}\|\|c_{t_2}\|}$. Then for each user $u$ we can compute the prediction for task $i$ as $r_{ui} = \sum_l sim(i,l)r_{il}$ where $l$ denotes the set of tasks user $u$ has performed.
% 
% \noindent {\bf Our Proposed Solutions:}
% We refer to our algorithms as follows: Feature Preference Model as {\tt Feature-Based-NNLS} and Implicit Factorization with Task Similarity as {\tt IFTS}. For the latter approach, we run 30 iterations of our algorithms. 

%{\bf Formally define precision and recall and provide references - what is Y and X suddenly - you need to be consistent with your notations}.
{\bf Evaluation Metrics:}
We use Mean Percentile Ranking(MPR) proposed by ~\cite{hu2008collaborative} for evaluating implicit feedback. The mathematical formula to calculate MPR is $\frac{\sum_{ij}c_{ij}\rho_{ij}}{\sum_{ij}c_{ij}}$. $\rho_{ij}$ is the percentile ranking of the task $j$ for worker $i$. Our recommendation is based on the estimated Worker-Task Preference matrix, $\hat{P}$. For {\tt Feat-Based-NNLS}, $\hat{P}= XY$, where $X$ is Worker-Feature matrix and $Y$ is Task-Feature matrix. For {\tt IFTS}, $\hat{P}= UV$. We experimented with different values of $\alpha$ and choose $\alpha$ = $50$. We also use Precision Recall curve as our second evaluation method. In this method, we want to evaluate our method based on how many task in the test set we can correctly predict by taking only (t\%) of the top-tasks. We vary $t$ (in an increment of $1\%$) in a continuous manner and  obtain PR curve.   

% \textbf{Mean Percentile Ranking(MPR)}: The mathematical formula to calculate MPR is $\frac{\sum_{ij}c_{ij}\rho_{ij}}{\sum_{ij}c_{ij}}$. Here, $c_{ij}$ indicates the number of task $t_j$ performed by worker $u_i$. $\rho_{ij}$ is the percentile ranking of the task $j$ for worker $i$. For instance, if a task $t_j$ is recommended as the first task, then it will be on the $0^{th}$ percentile, hence $\rho_{ij} =0$, or if it is the last task it is on the $100^{th}$ percentile, then $\rho_{ij} = 100$. If the tasks are recommended at random then the process has an expected MPR of $50$\%. MPR values become high, if the tasks completed by the worker higher number of times, are at the top of the sorted list. 
%This metric makes sense for our dataset as the mean of the throughput for the test set is [calculate ] and variance is [calculate], so it won't bias the score if a top ranked task has a very high throughput.
%
% \textbf{Precision-Recall(PR)Curve}: Precision is defined as the percentage of recommended tasks that are relevant, whereas, Recall means the percentage of relevant tasks that are retrieved. In this method, we want to evaluate our method based on how many task in the test set we can correctly predict by taking only (t\%) of the top-tasks. We vary $t$ (in an increment of $1\%$) in a continuous manner and  obtain PR curve.

%{\bf you need to provide intuition why one performs better or worse? - not just it performs better or worse} 

{\bf Summary of Results:}
The objective of our empirical study is to see how effective our proposed task recommendation models are in comparison with the baseline models. Our proposed algorithm {\tt Feature-Based-NNLS} convincingly outperforms the baseline algorithms in both MPR and PR-Curve. The reason behind the worse performance of {\tt Implicit-ALS-Negative} is that the worker does not choose tasks from a list of available task list, so a task that hasn't been attempted by the user really has no preference rather than ``negative preference''. {\tt IFTS} also performs reasonably well compare to other methods. 
\begin{figure}
\begin{floatrow}
\ffigbox{%
  \includegraphics[height=2.8cm, width=4.5cm] {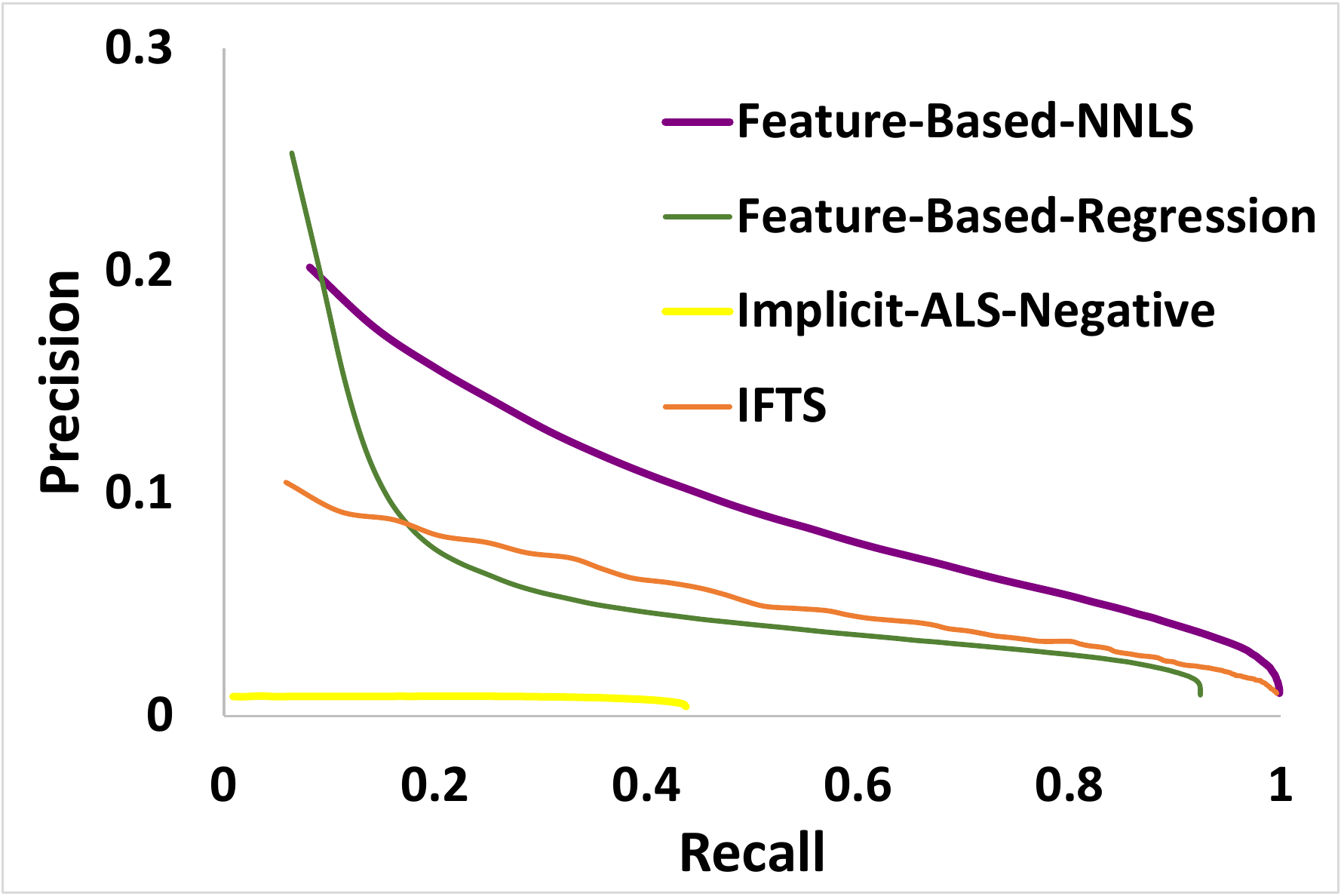}%
}{%
  \label{fig:PR}
  \caption{PR Curve}%
}
\capbtabbox{%
\begin{tabularx}{0.5\textwidth}{ b|s }
\hline
 Algorithm  & MPR\\ \hline
 {\tt Impl-ALS-Neg} & $17.3$ \\ \hline
 {\tt Feat-Based-Reg} & $13.706$  \\ \hline 
 {\bf {\tt Feat-Based-NNLS}} &  $\mathbf{5.68}$ \\ \hline
 {\tt IFTS} & $6.87$ \\
\hline
\end{tabularx}
}{%
  \label{tab:MPR}
  \caption{MPR}%
}
\end{floatrow}
\end{figure}

\vspace{-0.1in}
\section{Related Work}
Task recommendation with explicit observation is studied in~\cite{yuen2012taskrec}. We are the first to treat worker-task completion history as implicit observations and incorporate task feature information for recommendation. Works in recommender systems such as  ~\cite{forbes2011content,nguyen2013content,koren2008factorization} mostly rely on explicit feedback or content based feedback, whereas our model relies on implicit feedback. This precludes direct adaptation of their techniques.  
% 
% but  proposes classification based task recommender system, where the authors first create a user profile based on user meta-data using explicit feedback, then train a binary classifier to determine the likelihood of user selecting that particular task. Pick a crowd  uses social network as well as worker's information for task recommendation. 
%\vspace{-0.1in}
\section{Conclusion and Future Work} 
We initiate the study of the task recommendation problem in citizen science based crowdsourcing applications, considering both implicit feedback and explicit features. We formalize two optimization problems and present preliminary results. As ongoing research, we are investigating our method's validity on other datasets, as well as the generality of our proposed solution outside citizen science applications.
\newpage
%\vspace{-0.2in}

%In future work, we intend to present our results as well as detailed description of our proposed methods.              

%We design two elegant solutions that exploit implicit feedback, explicit features, as well as similarity between the tasks in constraining the latent factors of the tasks.  We formally analyze the complexity and present the efficacy of our proposed solutions by comparing against multiple state-of-the-art techniques.

 %and compare our quality of results with the current state of the art solution for task recommendation.

% As an ongoing research, we are  to look at the optimization problem fromr requester perspective and how it impacts the quality of recommendation.    

%\newpage
\bibliographystyle{aaai} 
\bibliography{references}

\begin{thebibliography}{}

\bibitem[\protect\citeauthoryear{Forbes and Zhu}{2011}]{forbes2011content}
Forbes, P., and Zhu, M.
\newblock 2011.
\newblock Content-boosted matrix factorization for recommender systems:
  experiments with recipe recommendation.
\newblock In {\em Proceedings of the fifth ACM conference on Recommender
  systems},  261--264.
\newblock ACM.

\bibitem[\protect\citeauthoryear{Geiger and
  others}{2014}]{geiger2014personalized}
Geiger, D., et~al.
\newblock 2014.
\newblock Personalized task recommendation in crowdsourcing information
  systems—current state of the art.
\newblock {\em Decision Support Systems} 65:3--16.

\bibitem[\protect\citeauthoryear{Hu and others}{2008}]{hu2008collaborative}
Hu, Y., et~al.
\newblock 2008.
\newblock Collaborative filtering for implicit feedback datasets.
\newblock In {\em ICDM}.

\bibitem[\protect\citeauthoryear{Koren}{2008}]{koren2008factorization}
Koren, Y.
\newblock 2008.
\newblock Factorization meets the neighborhood: a multifaceted collaborative
  filtering model.
\newblock In {\em Proceedings of the 14th ACM SIGKDD international conference
  on Knowledge discovery and data mining},  426--434.
\newblock ACM.

\bibitem[\protect\citeauthoryear{Lin and others}{2014}]{lin2014signals}
Lin, C.~H., et~al.
\newblock 2014.
\newblock Signals in the silence: Models of implicit feedback in a
  recommendation system for crowdsourcing.
\newblock In {\em AAAI}.

\bibitem[\protect\citeauthoryear{Nguyen and Zhu}{2013}]{nguyen2013content}
Nguyen, J., and Zhu, M.
\newblock 2013.
\newblock Content-boosted matrix factorization techniques for recommender
  systems.
\newblock {\em Statistical Analysis and Data Mining: The ASA Data Science
  Journal} 6(4):286--301.

\bibitem[\protect\citeauthoryear{Wu and others}{2006}]{wu2006learning}
Wu, Q., et~al.
\newblock 2006.
\newblock Learning rates of least-square regularized regression.
\newblock {\em Foundations of Computational Mathematics}.

\bibitem[\protect\citeauthoryear{Xue and others}{2013}]{xue2013improving}
Xue, Y., et~al.
\newblock 2013.
\newblock Improving your chances: Boosting citizen science discovery.
\newblock In {\em First AAAI Conference on Human Computation and
  Crowdsourcing}.

\bibitem[\protect\citeauthoryear{Yuen and others}{2012}]{yuen2012taskrec}
Yuen, M.-C., et~al.
\newblock 2012.
\newblock Taskrec: probabilistic matrix factorization in task recommendation in
  crowdsourcing systems.
\newblock In {\em Neural Information Processing}.

\end{thebibliography}
\end{document}